\documentclass[10pt,twocolumn,letterpaper]{article}
\usepackage{authblk}
\usepackage[pagenumbers]{cvpr}
\usepackage{graphicx}
\usepackage{amsmath}
\usepackage{amssymb}
\usepackage{booktabs}
\usepackage{array,multirow,graphicx}
\usepackage{algorithm}
\usepackage[noend]{algpseudocode}
\usepackage[toc,page]{appendix}
\usepackage{bbm}

\usepackage[pagebackref,breaklinks,colorlinks]{hyperref}

\usepackage[capitalize]{cleveref}
\crefname{section}{Sec.}{Secs.}
\Crefname{section}{Section}{Sections}
\Crefname{table}{Table}{Tables}
\crefname{table}{Tab.}{Tabs.}

\def\cvprPaperID{8351} 
\def\confName{CVPR}
\def\confYear{2022}

\begin{document}

\title{SegDiff: Image Segmentation with Diffusion Probabilistic Models}

\makeatletter
\newcommand\email[2][]%
   {\newaffiltrue\let\AB@blk@and\AB@pand
      \if\relax#1\relax\def\AB@note{\AB@thenote}\else\def\AB@note{\relax}%
        \setcounter{Maxaffil}{0}\fi
      \begingroup
        \let\protect\@unexpandable@protect
        \def\thanks{\protect\thanks}\def\footnote{\protect\footnote}%
        \@temptokena=\expandafter{\AB@authors}%
        {\def\\{\protect\\\protect\Affilfont}\xdef\AB@temp{#2}}%
         \xdef\AB@authors{\the\@temptokena\AB@las\AB@au@str
         \protect\\[\affilsep]\protect\Affilfont\AB@temp}%
         \gdef\AB@las{}\gdef\AB@au@str{}%
        {\def\\{, \ignorespaces}\xdef\AB@temp{#2}}%
        \@temptokena=\expandafter{\AB@affillist}%
        \xdef\AB@affillist{\the\@temptokena \AB@affilsep
          \AB@affilnote{}\protect\Affilfont\AB@temp}%
      \endgroup
       \let\AB@affilsep\AB@affilsepx
}
\makeatother

\author[1]{Tomer Amit} 
\author[1]{Tal Shaharbany}
\author[1,2]{Eliya Nachmani}
\author[1]{Lior Wolf}
\affil[1]{Tel-Aviv University} 
\affil[2]{Facebook AI Research}
\email{\url{{tomeramit1,shaharabany,eliyan,wolf}@mail.tau.ac.il}}

\maketitle

\begin{abstract}
   Diffusion Probabilistic Methods are employed for state-of-the-art image generation. In this work, we present a method for extending such models for performing image segmentation. The method learns end-to-end, without relying on a pre-trained backbone. The information in the input image and in the current estimation of the segmentation map is merged by summing the output of two encoders. Additional encoding layers and a decoder are then used to iteratively refine the segmentation map, using a diffusion model. Since the diffusion model is probabilistic, it is applied multiple times, and the results are merged into a final segmentation map. The new method produces state-of-the-art results on the Cityscapes validation set, the Vaihingen building segmentation benchmark, and the MoNuSeg dataset.
\end{abstract}

\section{Introduction}
\label{sec:intro}

Diffusion methods, which iteratively improve a given image, obtain image quality that is on par with or better than other types of generative models, including other forms of log-likelihood models and adversarial models~\cite{dhariwal2021diffusion,ho2022cascaded}. Such methods have been shown to excel in many generation tasks, both conditional and unconditional.

The vast majority of diffusion models are applied in domains in which there is no absolute ground truth result and the output is evaluated either through a user study or using several quality and diversity scores. As far as we know, with the exception of super resolution~\cite{ho2022cascaded,saharia2021image,li2022srdiff}, diffusion models have not been applied to problems in which the ground truth result is unique.

In this work, we tackle the problem of image segmentation. This problem is a cornerstone of both classical computer vision and the deep learning methods of the last decade. The leading methods in the field employ encoder-decoder networks of varied architectures~\cite{long2015fully,ronneberger2015u,xie2021segformer,zhou2018unet++,chao2019hardnet,yu2015multi}. While adversarial methods have been attempted~\cite{luc2016semantic,xie2017adversarial,xue2018segan,fischer2017adversarial}, they do not constitute the current state of the art.

Therefore, it is uncertain whether diffusion models, which have been used primarily for GAN-like generation tasks, would be competitive in this domain. In this work, we propose applying a diffusion model to learn the image segmentation map. Unlike other recent improvements in the field of image segmentation~\cite{strudel2021segmenter,fu2020scene,huang2019ccnet}, we train our method end-to-end, without relying on a pre-trained backbone network.

The diffusion model employs a denoising network conditioned on the input image only through a sum in which this information is aggregated with information arising from the current estimate $x_t$. Specifically, the input image $I$ and the current estimate $x_t$ of the binary segmentation map are passed through two different encoders, and the sum of these multi-channel tensors is passed through a U-Net~\cite{ronneberger2015u} to provide the next estimate $x_{t-1}$.

Since the generation process is stochastic in its nature, one may obtain multiple solutions. As we show, merging these solutions, by simply averaging multiple runs, leads to an improvement in overall accuracy.

The novel method presented produces state-of-the-art results on multiple benchmarks: Cityscapes~\cite{Cordts2016Cityscapes}, building segmentation~\cite{rottensteiner2014isprs}, and nuclei segmentation~\cite{kumar2019multi,kumar2017dataset}. 

Our main contributions are:
\begin{itemize}
  \item We are the first to apply diffusion models to the image segmentation problem. 
  \item We propose a new way to condition the model on the input image.
  \item We introduce the concept of multiple generations, in order to improve performance and calibration of the diffusion model.
  \item We obtained state-of-the-art results on multiple benchmarks. The margin is especially large for small data sets.
\end{itemize}

\begin{figure*}[t]
    \centering
    \includegraphics[width=1.00\linewidth]{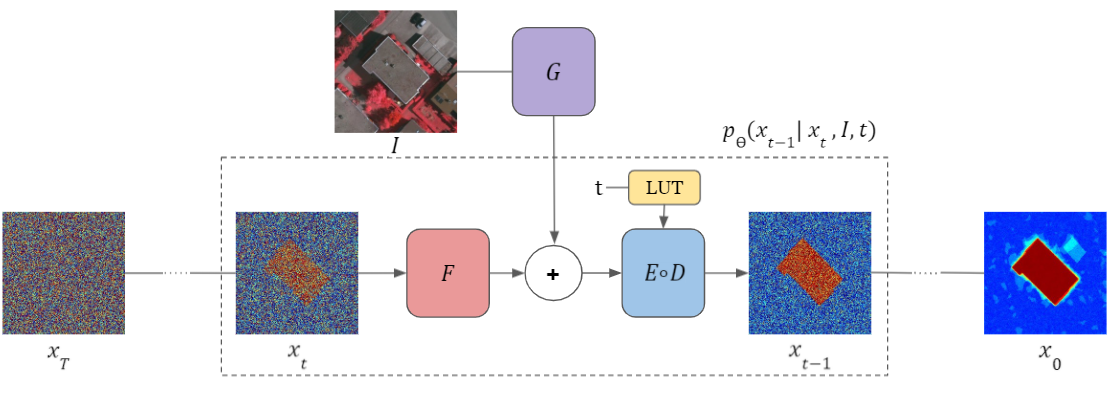}
    \caption{Our proposed diffusion method for image segmentation encodes the input signal, $x_t$, with $F$. The extracted features are summed with the feature map of the conditioned image $I$ generated by network $G$. Networks $E$ and $D$ are a U-net encoder and decoder~\cite{ronneberger2015u,nichol2021improved}, respectively, that refine the estimated segmentation map, obtaining $x_{t-1}$.}
    \label{fig:method}
\end{figure*}

\section{Related work}
\noindent{\bf Image segmentation\quad} is a problem of assigning each pixel a label that identifies whether it belongs to a specific class or not. This problem is widely investigated using different architectures. These include fully convolutional networks~\cite{long2015fully}, encoder-decoder architectures with skip-connections, such as U-Net~\cite{ronneberger2015u}, transformer-based architectures, such as the segformer~\cite{xie2021segformer}, and even architectures that combine hypernetworks, such as~\cite{nirkin2021hyperseg}.

\noindent{\bf Diffusion Probabilistic Models (DPM)~\cite{sohl2015deep}\quad} are a class of generative models based on a Markov chain, which can transform a simple distribution (e.g. Gaussian) to data that is sampled in a complex distribution. Diffusion models are capable of generating high-quality images that can compete with and even outperform the latest GAN methods~\cite{sohl2015deep,ho2020denoising,nichol2021improved,dhariwal2021diffusion}. A variational framework for the likelihood estimation of diffusion models was introduced by Huang et al.~\cite{huang2021variational}. Subsequently, Kingma et al.~\cite{kingma2021variational} proposed a Variational Diffusion Model that produces state-of-the-art results in likelihood estimation for image density. 

Diffusion models were also applied to language modeling \cite{hoogeboom2021argmax,austin2021structured}, where a novel diffusion model for categorical data was used.

\noindent{\bf Conditional Diffusion Probabilistic Models\quad} In our work, we use diffusion models to solve the image segmentation problem as conditional generation, given the image. Conditional generation with diffusion models includes methods for class-conditioned generation, which is obtained by adding a class embedding to the timestamp embedding~\cite{nichol2021improved}. In \cite{choi2021ilvr} a method for guiding the generative process in DDPM is present. This method allows the generation of images based on a given reference image without any additional learning. 

In the domain of super resolution, the lower-resolution image is upsampled and then concatenated, channelwise, to the generated image at each iteration~\cite{saharia2021image,ho2022cascaded}. A similar approach passes the low-resolution images through a convolutional block~\cite{li2022srdiff} prior to the concatenation. Concurrently with our work, diffusion models were applied to image-to-image translation tasks~\cite{saharia2021palette}. These tasks include uncropping, inpainting, and colorization. The results obtained outperform strong GAN baselines.

Conditional diffusion models have also been used for voice generation. The mel-spectrogram is processed with a convolutional network, and is used as an additional input to the DPM denoising network $\epsilon$~\cite{chen2020wavegrad,kong2020diffwave,liu2021diffsinger}. Furthermore, in \cite{popov2021grad} a text-to-speech diffusion model is introduced, which uses text as a condition to the diffusion model.

In our work, we take a different approach to conditioning, adding (not concatenating) the input image, after it passes through an convolutional encoder, to the current estimation of the segmentation image. In other words, we learn the DPM of a residual model.


\section{Background}

We  briefly introduce the formulation of diffusion models mentioned in~\cite{ho2020denoising}.
Diffusion models are generative models parametrized by a Markov chain and composed of forward and backward processes. The forward process q is described by the formulation:
\begin{equation}
    q(x_{1:T}|x_0) 
    = 
    \prod^T_{t=1}q(x_t|x_{t-1}),
\end{equation}
where T is the number of steps in the diffusion model, $x_1, ... , x_T$ are latent variables, and $x_0$ is a sample from the data. At each iteration of the forward process, Gaussian noise is added according to
\begin{equation}
    q(x_t|x_{t-1}) 
    = 
    N(x_t;\sqrt{1-\beta_t}x_{t-1},\beta_t I_{n\times n}),
\end{equation}
where $\beta_t$ is a constant that defines the schedule of added noise, and $I_{n\times n}$ is the identity matrix of size n. As described in~\cite{ho2020denoising}, 
\begin{equation}
    \alpha_t 
    = 
    1-\beta_t, 
    \Bar{\alpha}_t
    = 
    \prod^t_{s=0}\alpha_s.
\end{equation}

The forward process supports sampling at an arbitrary timestamp $t$, with the formula
\begin{equation}
    q(x_t|x_0) 
    = 
    N(x_t;\sqrt{\Bar{\alpha}_t}x_0,(1-\Bar{\alpha}_t)I_{n\times n}),
\end{equation}
which can be reparametrized to:
\begin{equation}
    x_t 
    = 
    \sqrt{\Bar{\alpha}_t}x_0 + \sqrt{(1-\Bar{\alpha}_t)} \epsilon, \epsilon \thicksim N(0,I_{n\times n}).
\label{eq:xt_reparametrization}
\end{equation}
The reverse process is parametrized by $\theta$ and defined by
\begin{equation}
    p_\theta(x_{0:T-1}|x_T) 
    = 
    \prod^T_{t=1}p_\theta(x_{t-1}|x_t).
\end{equation}

Starting from $p_\theta(x_T)=N(x_T;0,I_{n\times n})$, the reverse process transforms the latent variable distribution $p_\theta(x_T)$ to the data distribution $p_\theta(x_0)$. The reverse process steps are performed by taking small Gaussian steps described by 
\begin{equation}
    p_\theta(x_{t-1}|x_t) 
    = 
    N(x_{t-1};\mu_\theta(x_t,t), \Sigma_\theta(x_t,t)).
\label{eq:reverse_process}
\end{equation}

Calculating $q(x_{t-1}|x_t,x_0)$ using Bayes' theorem, one obtains:
\begin{equation}
    q(x_{t-1}|x_t,x_0) 
    = 
    N(x_{t-1}; \tilde{\mu}(x_t, x_0), \tilde{\beta}_t I_{n\times n}),
\end{equation}
where 
\begin{equation}
    \tilde{\mu}_t(x_t, x_0) 
    = 
    \frac{\sqrt{\Bar{\alpha}_{t-1}}\beta_t}{1-\Bar{\alpha}_t}x_0 + \frac{\sqrt{\alpha_t}(1-\Bar{\alpha}_{t-1})}{1-\Bar{\alpha}_t}x_t,
\label{eq:mu_tilda}
\end{equation}
\begin{equation}
    \tilde{\beta} 
    = 
    \frac{1-\Bar{\alpha}_{t-1}}{1-\Bar{\alpha}_t}\beta_t.
\end{equation}
The neural network $\mu_\theta$ predicts the noise $\epsilon$, which is parametrized using Eq.~\ref{eq:xt_reparametrization},~\ref{eq:mu_tilda} to obtain:
\begin{equation}
    \mu_\theta(x_t,t)
    =
    \frac{1}{\sqrt{\alpha_t}}(x_t - \frac{\beta_t}{\sqrt{1-\Bar{\alpha}_t}}\epsilon_\theta(x_t,t)).
\label{eq:mu_theta}
\end{equation}
Following~\cite{ho2020denoising} we set 
\begin{equation}
    \Sigma_\theta(x_t,t)
    =
    \sigma^2_t I_{n\times n},
\end{equation}
where  
\begin{equation}
    \sigma^2_t
    =
    \tilde{\beta}_t.
\end{equation}
The forward process variance parameter is chosen to be a linearly increasing constant from $\beta_1 = 10^{-4}$ to $\beta_T = 2*10^{-2}$, formally:
\begin{equation}
    \beta_t
    =
    \frac{10^{-4}(T-t) + 2*10^{-2}(t-1)}{T-1}.
\label{eq:beta_t}
\end{equation}
Finally, we will minimize the term 
\begin{equation}
    E_{x_0,\epsilon,t}[||\epsilon-\epsilon_\theta(\sqrt{\Bar{\alpha}_t}x_0 + \sqrt{1-\Bar{\alpha}_t}\epsilon, t)||^2],
\label{eq:expectation_loss_term}
\end{equation}
where $\epsilon \thicksim N(0,I_{n\times n})$.

For inference, we can reparametrize the reverse process, Eq.~\ref{eq:reverse_process}, with Eq.~\ref{eq:mu_theta}, obtaining
\begin{equation}
    x_{t-1} 
    = 
    \frac{1}{\sqrt{\alpha_t}}(x_t - \frac{1-\alpha_t}{\sqrt{1-\Bar{\alpha}_t}}\epsilon_\theta(x_t,t)) + \sigma_\theta z.
\label{eq:reverse_process_rep}
\end{equation}

\begin{table}[t]
\begin{minipage}[t]{0.50\textwidth}
\begin{algorithm}[H]
    \caption{Inference Algorithm}
    \begin{algorithmic}
        \State \textbf{Input} total diffusion steps T, image $I$
        \State $x_T \thicksim N(\mathbf{0},\mathbf{I_{n\times n}})$
        \For{$t=T,T-1,...,1$}
                \State $z \thicksim N(\mathbf{0},\mathbf{I_{n\times n}})$
            \State $\beta_t=\frac{10^{-4}(T-t) + 2*10^{-2}(t-1)}{T-1}$
            \State $\alpha_t  = 1-\beta_t$
            \State $\Bar{\alpha}_t = \prod^t_{s=0}\alpha_s$
            \State $\tilde{\beta}_t = \frac{1-\Bar{\alpha}_{t-1}}{1-\Bar{\alpha}_t}\beta_t$
            \State $x_{t-1} = {\alpha_t}^{-\scriptscriptstyle\frac{1}{2}}(x_t - \frac{1-\alpha_t}{\sqrt{1-\Bar{\alpha}_t}}\epsilon_\theta(x_t,I,t)) +  \mathbbm{1}_{[t > 1]}{\tilde{\beta}_t}^{\scriptscriptstyle\frac{1}{2}} z$
        \EndFor
        \State \Return $x_0$
    \end{algorithmic}
\label{alg:Inference}
\end{algorithm}
\end{minipage}

\begin{minipage}[t]{0.46\textwidth}
\begin{algorithm}[H]
    \caption{Training Algorithm}
    \begin{algorithmic}
        \State \textbf{Input} total diffusion steps T, images and segmentation masks dataset $D = \{(I_k,M_k)\}^K_k=1$
        \Repeat
            \State Sample $(I_i,M_i) \thicksim D$, $\epsilon \thicksim N(\mathbf{0},\mathbf{I_{n\times n}})$
            \State Sample $t \thicksim $ Uniform(\{1,...,T\})
            \State $\beta_t=\frac{10^{-4}(T-t) + 2*10^{-2}(t-1)}{T-1}$
            \State $\alpha_t  = 1-\beta_t$
            \State $\Bar{\alpha}_t = \prod^t_{s=0}\alpha_s$
            \State Take gradient step on $\nabla_\theta|| \epsilon - \epsilon_\theta (x_t,I_i,t)||, x_t=\sqrt{\Bar{\alpha_t}}M_i + \sqrt{1-\Bar{\alpha_t}}\epsilon$
        \Until{convergence}
    \end{algorithmic}
\label{alg:Training}
\end{algorithm}
\end{minipage}
\end{table}

\section{Method}

Our method modifies the diffusion model by conditioning the step estimation function $\epsilon_\theta$ on an input tensor that combines information derived from both the current estimate $x_t$ and the input image $I$.

In diffusion models, $\epsilon_\theta$ is typically a U-Net~\cite{ronneberger2015u}. In our work, $\epsilon_\theta$ can be expressed in the following form:
\begin{equation}
\label{eq:newepsilon}
    \epsilon_\theta (x_t,I,t) = D(E(F(x_t)+G(I),t),t)\,.
\end{equation}
In this architecture, the U-Net's decoder $D$ is conventional and its encoder is broken down into three networks: $E$, $F$, and $G$. The last encodes the input image, while $F$ encodes the segmentation map of the current step $x_t$. The two processed inputs have the same spatial dimensionality and number of channels. Based on the success of residual connections~\cite{he2016deep}, we sum these signals $F(x_t)+G(I)$. This sum then passes to the rest of the U-Net encoder $E$.

The current step index $t$ is passed to two different networks $D$ and $E$. In each of these, it is embedded using a shared learned look-up table. 

The output of $\epsilon_\theta$ from Eq.~\ref{eq:newepsilon}, which is conditioned on $I$, is plugged into Eq.~\ref{eq:reverse_process_rep}, replacing the unconditioned $\epsilon_\theta$ network. This resulting inference time procedure is illustrated in Fig.~\ref{fig:method} and detailed in Alg.~\ref{alg:Inference}.

 \begin{figure*}[t]
\centering
\begin{tabular}{@{}c@{~}c@{~}c@{~}c@{~}c@{~}c@{~}c@{}}
    \includegraphics[width=0.16135\textwidth]{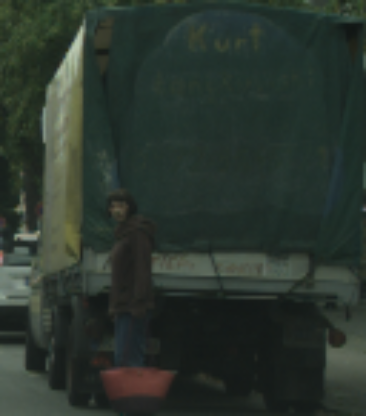} &   \includegraphics[width=0.16135\textwidth]{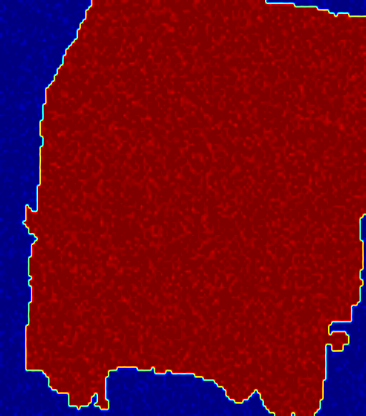} &  
    \includegraphics[width=0.16135\textwidth]{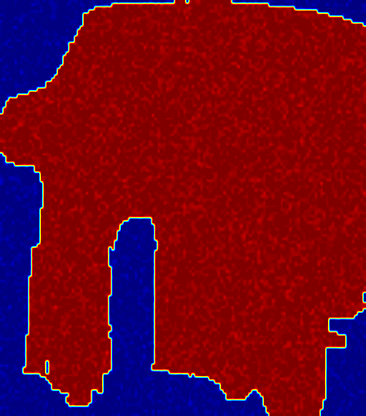} & 
    \includegraphics[width=0.16135\textwidth]{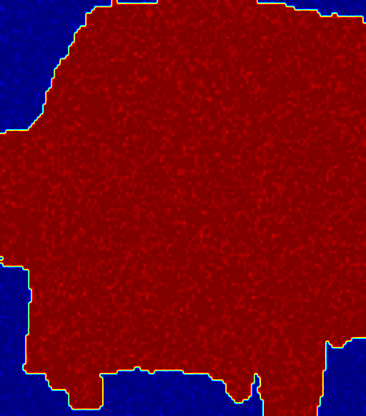}&
    \includegraphics[width=0.16135\textwidth]{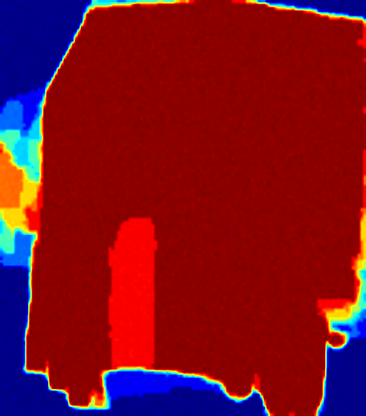} &  
     \includegraphics[width=0.16135\textwidth]{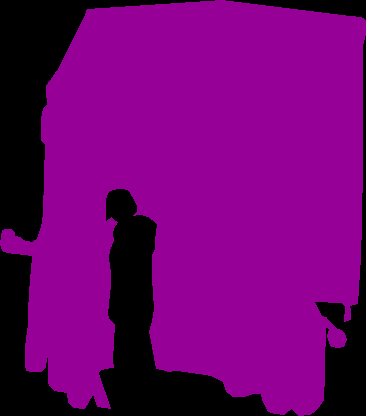}\\
        \includegraphics[width=0.16135\textwidth]{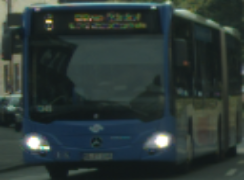} &   \includegraphics[width=0.16135\textwidth]{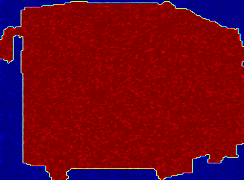} &  \includegraphics[width=0.16135\textwidth]{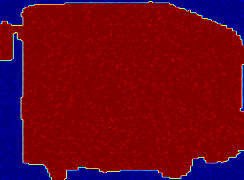} & 
        \includegraphics[width=0.16135\textwidth]{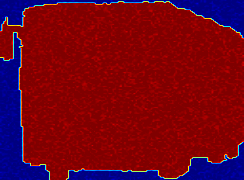} & \includegraphics[width=0.16135\textwidth]{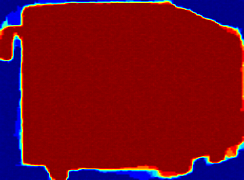} &    \includegraphics[width=0.16135\textwidth]{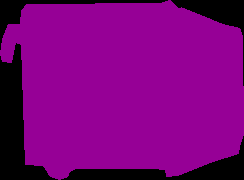}\\
        
        \includegraphics[width=0.16135\textwidth]{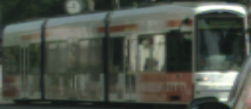} &   \includegraphics[width=0.16135\textwidth]{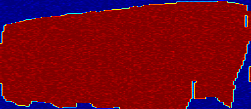} &  \includegraphics[width=0.16135\textwidth]{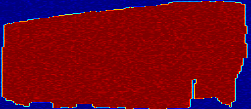} & 
        \includegraphics[width=0.16135\textwidth]{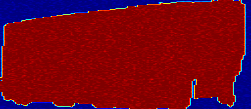} & \includegraphics[width=0.16135\textwidth]{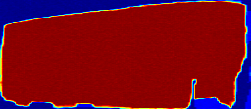} &   \includegraphics[width=0.16135\textwidth]{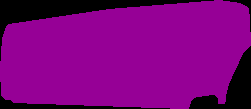}\\
        
        \includegraphics[width=0.16135\textwidth]{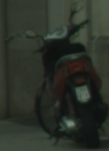} &   \includegraphics[width=0.16135\textwidth]{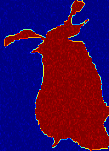} & \includegraphics[width=0.16135\textwidth]{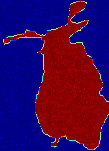} & \includegraphics[width=0.16135\textwidth]{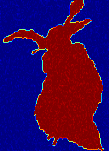} & \includegraphics[width=0.16135\textwidth]{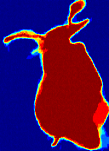} &  \includegraphics[width=0.16135\textwidth]{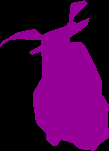}\\

        \includegraphics[width=0.16135\textwidth]{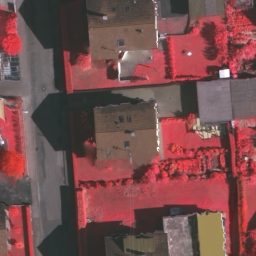} &   \includegraphics[width=0.16135\textwidth]{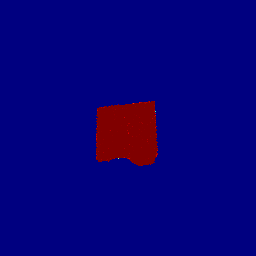} & \includegraphics[width=0.16135\textwidth]{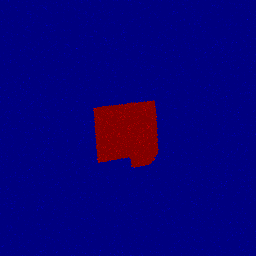} & \includegraphics[width=0.16135\textwidth]{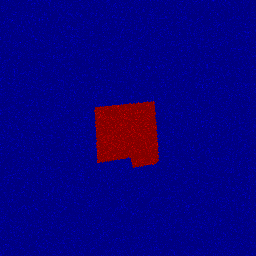} & \includegraphics[width=0.16135\textwidth]{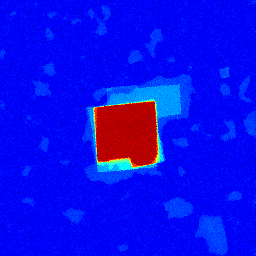} &  \includegraphics[width=0.16135\textwidth]{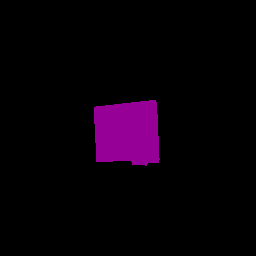}\\    
        
        \includegraphics[width=0.16135\textwidth]{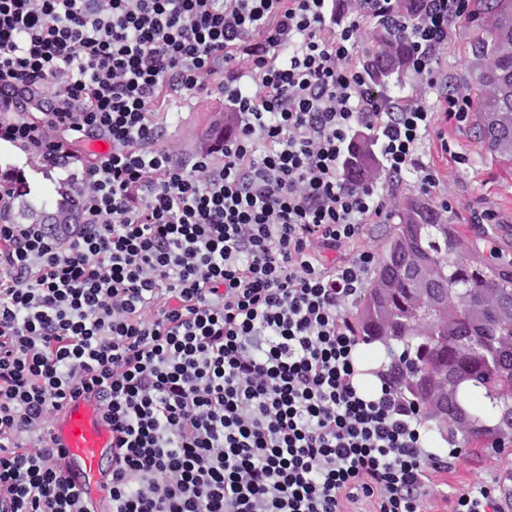} &   \includegraphics[width=0.16135\textwidth]{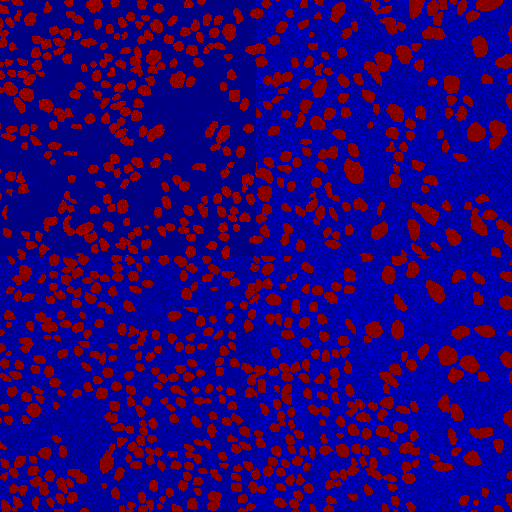} & \includegraphics[width=0.16135\textwidth]{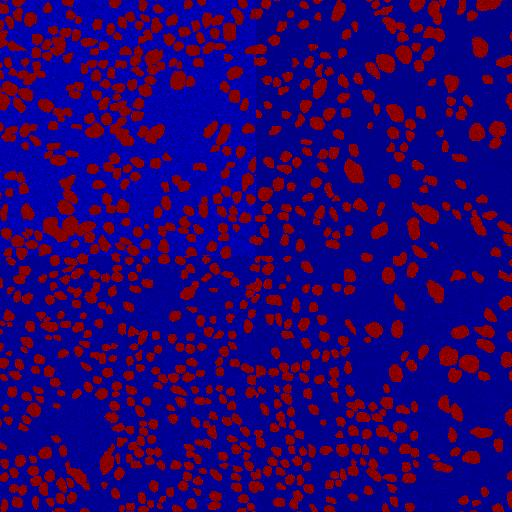} & \includegraphics[width=0.16135\textwidth]{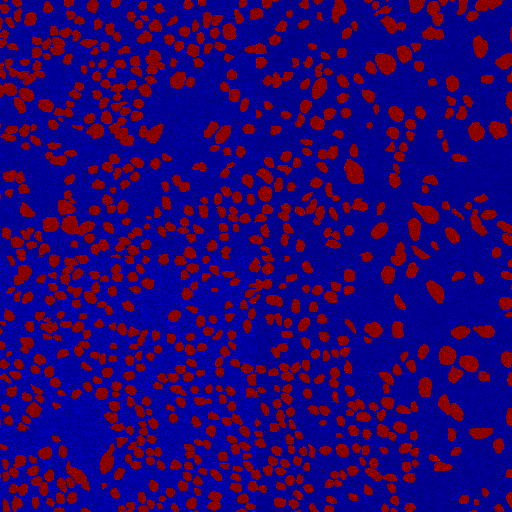} & \includegraphics[width=0.16135\textwidth]{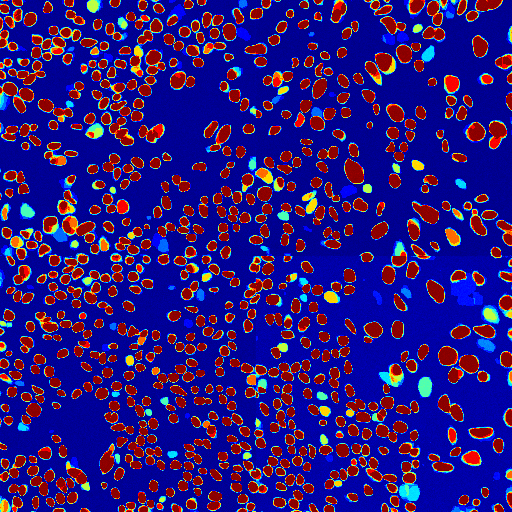} &  \includegraphics[width=0.16135\textwidth]{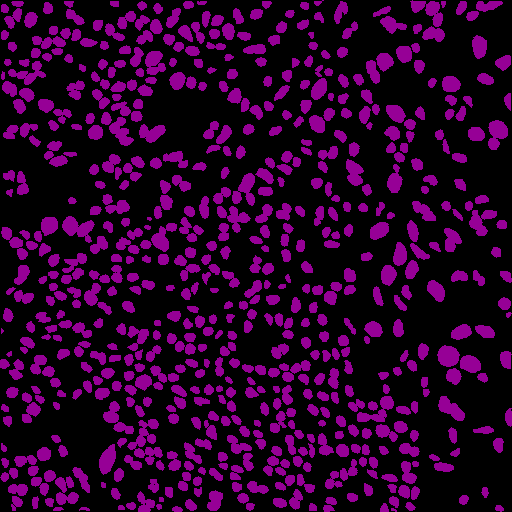}\\
        
           \cmidrule(lr){1-1} \cmidrule(lr){2-4}\cmidrule(lr){5-5}\cmidrule(lr){6-6}
            (a) & \multicolumn{3}{c}{(b)} & (c) & (d)\\
            
\end{tabular}
\caption{Obtaining multiple segmentation results for Cityscapes, Vaihingen, and MoNuSeg. (a) input image, (b) a subset of the obtained results for multiple runs on the same input, visualized by the jet color scale between 0 in blue and 1 in red, (c) average result, and (d) ground truth.}
\label{fig:variabiltiy}
\end{figure*}

\subsection{Employing multiple generations}

Since calculating $x_{t-1}$ during inference includes the addition of $\sigma_\theta(x_t,t)z$, where $z$ is from a standard distribution, there is significant variability between different runs of the inference method on the same inputs, see Fig.~\ref{fig:variabiltiy}(b). 

In order to exploit this phenomenon, we run the inference algorithm multiple times, then average the results. 

This way, we stabilize the results of segmentation and improve performance, as demonstrated in Fig.~\ref{fig:variabiltiy}(c). We use thirty generated instances in all experiments, except for the experiments in the ablation study, which quantifies the gain of this averaging procedure.

\subsection{Training}

The training procedure is depicted in Alg.~\ref{alg:Training}. The total number of diffusion steps $T$ is set by the user. For each iteration, a random sample is obtained $(I_i,M_i)$ (an image and the associated ground truth binary segmentation map). The iteration number $1\leq t\leq T$ is sampled from a uniform distribution, and epsilon from a standard distribution. 

We then sample $x_t$ according to Eq.~\ref{eq:xt_reparametrization}, compute $F(x_t)+G(I_i)$, and apply networks $E$ and $D$ to obtain $\epsilon_\theta(x_t,I_i,t)$.

The loss being minimized is a modified version of Eq~\ref{eq:expectation_loss_term}, namely:
\begin{equation}
    E_{x_0,\epsilon,x_e,t}[||\epsilon-\epsilon_\theta(\sqrt{\Bar{\alpha}_t}x_0 + \sqrt{1-\Bar{\alpha}_t}\epsilon, I_i, t)||^2].
\label{eq:conditional_loss_term}
\end{equation}

At training time, the ground truth segmentation of the input image $I_i$ is known, and the loss is computed by setting $x_0 = M_i$.

\subsection{Architecture}

The input image encoder $G$ is built from Residual in Residual Dense Blocks~\cite{wang2018esrgan} (RRDBs), which combine multi-level residual connections without batch normalization layers. $G$ has an input 2D-convolutional layer, an RRDB with a residual connection around it, followed by another 2D-convolutional layer, leaky RELU activation and a final 2D-convolutional output layer. $F$ is a 2D-convolutional layer with a single-channel input and an output of $C$ channels.

The encoder-decoder part of $\epsilon_\theta$, i.e., $D$ and $E$, is based on U-Net, similarly to~\cite{nichol2021improved}. Each level is composed of residual blocks, and at resolution 16x16 and 8x8 each residual block is followed by an attention layer. The bottleneck contains two residual blocks with an attention layer in between. Each attention layer contains multiple attention heads.

The residual block is composed of two convolutional blocks, where each convolutional block contains group-norm, Silu activation, and a 2D-convolutional layer. The residual block receives the time embedding through a linear layer, Silu activation, and another linear layer. The result is then added to the output of the first 2D-convolutional block. Additionally, the residual block has a residual connection that passes all its content.

On the encoder side (network $E$), there is a downsample block after the residual blocks of the same depth, which is a 2D-convolutional layer with a stride of two. On the decoder side (network $D$), there is an upsample block after the residual blocks of the same depth, which is composed of the nearest interpolation that doubles the spatial size, followed by a 2D-convolutional layer. Each layer in the encoder has a skip connection to the decoder side.

\section{Experiments}
\label{sec:experiments}

We present segmentation results for three datasets, as well as an ablation study. 

\noindent{\bf Datasets\quad} The Cityscapes dataset~\cite{Cordts2016Cityscapes} is an instance segmentation dataset containing 5,000 annotated images divided into 2,975 images for training, 500 for validation, and 1,525 for testing. 

The experimental setting used is sometimes referred to as interactive segmentation and is motivated by the need to accelerate object annotation~\cite{acuna2018efficient}. Under this setting, there are eight object categories, and the goal is to recover the objects' per-pixel masks, given a cropped patch that contains the bounding box around each object.

Our per-object training and validation sets are created by taking crops from images in the original Cityscapes sets using the locations of the ground truth classes (we do not have access to the ground truth labels of the original Cityscapes test set).

We compared our method for the Cityscapes dataset with PSPDeepLab~\cite{chen2017deeplab},  Polygon-RNN++~\cite{acuna2018efficient},  Curve-GCN~\cite{ling2019fast} Deep active contours~\cite{gur2019end}, Segformer-B5~\cite{xie2021segformer} and Stdc1~\cite{fan2021rethinking}. For most baselines, we report the results obtained from previous publications. For Segformer and Stdc, we train from scratch. 

We did not perform a comparison with PolyTransform~\cite{liang2020polytransform}, since it uses a different protocol. Specifically, this method, which improves upon Mask R-CNN~\cite{he2017mask}, utilizes the entire image (and not just the segmentation patch) as part of its inputs, and does not work on standard patches in a way that would enable a direct comparison.

The Vaihingen dataset~\cite{rottensteiner2014isprs} contains 168 aerial images of Vaihingen, in Germany, divided into 100 images for training and 68 for the test. The task is to segment the central building in each image. For this dataset, the leading baselines are DSAC~\cite{marcos2018learning}, DarNet~\cite{cheng2019darnet}, TDAC~\cite{hatamizadeh2020end}, Deep active contours~\cite{gur2019end}, FCN-UNET~\cite{ronneberger2015u}, FCN-ResNet-34, FCN-HarDNet-85~\cite{chao2019hardnet}, Segformer-B5~\cite{xie2021segformer} and Stdc1~\cite{fan2021rethinking}.

The MoNuSeg dataset~\cite{kumar2019multi,kumar2017dataset} contains a training set with 30 microscopic images from seven organs, with annotations of 21,623 individual nuclei. The test dataset contains 14 similar images. We resized the images to a resolution of $512\times 512$, following~\cite{valanarasu2021medical}. The relevant baseline methods are FCN~\cite{badrinarayanan2017segnet}, UNET~\cite{ronneberger2015u},  UNET++~\cite{zhou2018unet++}, Res-Unet~\cite{xiao2018weighted}, Axial attention (A.A) Unet~\cite{wang2020axial} and Medical transformer~\cite{valanarasu2021medical}.

\noindent{\bf Evaluation\quad}
The Cityscapes dataset is evaluated using the common metrics of mean Intersection-over-Union (mIoU) per class.
\begin{equation}
mIoU(y_i,\hat{y_i}) =  \sum_{i=1}^N \frac{TP(y_i,\hat{y_i})}{TP(y_i,\hat{y_i}) + FN(y_i,\hat{y_i}) + FP(y_i,\hat{y_i})}
\end{equation}
Where $N$ is the number of classes in the dataset, TP is the true positive between the ground truth $y$ and output mask $\hat{y}$, FN is a false negative, and FP is a false positive.

The Vaihingen dataset is evaluated using several metrics: mIoU, F1-score, Weighted Coverage (WCov), and Boundary F-score (BoundF), as described in~\cite{cheng2019darnet}. Briefly, the prediction is
correct if it is within a certain distance threshold from the
ground truth. The benchmarks use five thresholds, from 1px to 5px, for evaluating performance.

Following previous work, evaluation on the MoNuSeg dataset is performed using mIoU and the F1-score.

\noindent{\bf Training details\quad}
The number of diffusion steps in previous works was 1000~\cite{ho2020denoising} and even 4000~\cite{nichol2021improved}. The literature suggests that more is better~\cite{san2021noise}. In our main experiments, we employ 100 diffusion steps to reduce inference time. An additional set of experiments investigated the influence of the number of diffusion steps on the performance and runtime of the method.

The  AdamW~\cite{loshchilov2017decoupled} optimizer is used in all our experiments. Based on the intuition that the more RRDB blocks, the better the results, we used as many blocks as we could fit on the GPU without overly reducing batch size. The Unet used for datasets with a resolution of 256 $\times$ 256 has one additional layer with respect to the dataset with half that resolution, in order to account for the spatial dimensions.

On the Cityscapes dataset, the input resolution of our model is $128 \times 128$. The test metrics are computed on the original resolution; therefore, we resized the prediction to the original image size. 

Training took place with a batch size of 30 images. The network had 15 RRDB blocks and a depth of six. The number of channels was set to $[C, C, 2C, 2C, 4C, 4C]$ with $C=128$. 
We followed the same augmentation scheme as in~\cite{gur2019end}, including random scaling in the range of $[0.75, 1.25]$, with up to 22 degrees rotation in each direction, and a horizontal flip with a probability of 0.5.

For the Vaihingen dataset, the size of the input image and the test image resolution was $256 \times 256$. The experiments were performed with a batch size of eight images, six RRDB blocks, and a depth of seven. The number of channels was set to $[C, C, C, 2C, 2C, 4C, 4C]$ with $C=128$. 

The same augmentations are used as in~\cite{cheng2019darnet}: random scaling by a factor sampled uniformly in the range $[0.75, 1.5]$, a rotation sampled uniformly between zero and 360 degrees, independent horizontal and vertical flips, applied with a probability of 0.5, and a random color jitter, with a maximum value of 0.6 brightness, 0.5 contrast, 0.4 saturation, and 0.025 hue.

For MoNuSeg, the input image resolution was $256 \times 256$, but the test resolution was $512 \times 512$. To address this, we applied a sliding window of $256 \times 256$ with a stride of $256$, i.e., we tested each quadrant of the image separately. 

\begin{table*}[t]
\centering
\hspace{-0.2cm}
 \begin{tabular}{c l c c c c c c c c c }
 \midrule
 & \multicolumn{1}{l}{Method} & \multicolumn{1}{c}{Bicycle} & \multicolumn{1}{c}{Bus} & \multicolumn{1}{c}{Person} & \multicolumn{1}{c}{Train} & \multicolumn{1}{c}{Truck} & \multicolumn{1}{c}{M.cycle} & \multicolumn{1}{c}{Car} &  \multicolumn{1}{c}{Rider} & \multicolumn{1}{c}{Mean}\\
 \midrule
 \parbox[t]{3mm}{\multirow{5}{*}{\rotatebox[origin=c]{90}{expansion}}} 
& Polygon-RNN++~\cite{acuna2018efficient}
& 63.06 & 81.38 & 72.41 & 64.28 & 78.90 & 62.01 & 79.08 & 69.95 & 71.38 \\
& PSP-DeepLab~\cite{chen2017deeplab}
& 67.18 & 83.81 & 72.62 & 68.76 & 80.48 & 65.94 & 80.45 & 70.00 & 73.66 \\
& Polygon-GCN~\cite{ling2019fast}
& 66.55 & 85.01 & 72.94 & 60.99 & 79.78 & 63.87 & 81.09 & 71.00 & 72.66 \\
& Spline-GCN~\cite{ling2019fast}
& 67.36 & 85.43 & 73.72 & 64.40 & 80.22 & 64.86 & 81.88 & 71.73 & 73.70 \\
& SegDiff (ours)
& \textbf{69.80} & \textbf{85.97} & \textbf{76.09} & \textbf{75.95} & \textbf{80.68} & \textbf{67.06} & \textbf{83.40} & \textbf{72.57} & \textbf{76.44} \\
\midrule
\midrule
& Deep contour~\cite{gur2019end}
& 68.08 & 83.02 & 75.04 & 74.53 & 79.55 & 66.53 & 81.92 & 72.03 & 75.09 \\
& Segformer-B5~\cite{xie2021segformer}
& 68.02 & 78.78 & 73.53 & 68.46 & 74.54 & 64.06 & 83.20 & 69.12 & 72.46 \\
& Stdc1~\cite{fan2021rethinking}
& 67.86 & 80.67 & 74.20 & 69.73 & 77.02 & 64.52 & 83.53 & 69.58 & 73.39 \\
& Stdc2~\cite{fan2021rethinking}
& 68.67 & 81.29 & 74.41 & 71.36 & 75.71 & 63.69 & 83.51 & 69.90 & 73.57 \\
& SegDiff (ours)
& \textbf{69.62} & \textbf{84.64} & \textbf{75.18} & \textbf{74.89} & \textbf{80.34} & \textbf{67.75} & \textbf{83.63} & \textbf{73.49} & \textbf{76.19} \\
\midrule
 \end{tabular}
\captionof{table}{Cityscapes segmentation results for two protocols: the top part refers to segmentation results with 15\% expansion
around the bounding box; the bottom part refers to segmentation results with a tight bounding box.}
\label{tab:cityscapes}
\vspace{-3mm}
\end{table*}

\begin{table}[t]
\centering
\begin{tabular}{@{}l@{~}cccc@{}}
\hline
Method & F1-Score & mIoU & WCov & FBound \\
\hline
FCN-UNet~\cite{ronneberger2015u}
& 87.40 & 78.60 & 81.80 & 40.20 \\
FCN-ResNet34
& 91.76 & 87.20 & 88.55 & 75.12\\
FCN-HarDNet~\cite{chao2019hardnet}
& 93.97 & 88.95 & 93.60 & 80.20\\
DSAC~\cite{marcos2018learning}
& - & 71.10 & 70.70 & 36.40 \\
DarNet~\cite{cheng2019darnet}
& 93.66 & 88.20 & 88.10 & 75.90 \\
Deep contour~\cite{gur2019end}
& 94.80 & 90.33 & 93.72 & 78.72 \\
TDAC~\cite{hatamizadeh2020end}
& 94.26 & 89.16 & 90.54 & 78.12 \\
Segformer-B5~\cite{xie2021segformer}
& 93.94 & 88.57 & 91.91 & 77.95 \\
Stdc1~\cite{fan2021rethinking}
& 94.04 & 88.75 & 92.78 & 78.86 \\
Stdc2~\cite{fan2021rethinking}
& 93.97 & 88.62 & 92.59 & 77.3 \\
SegDiff (ours)
& \textbf{95.14} & \textbf{91.12} & \textbf{93.83} & \textbf{85.09}\\
\hline
\end{tabular}
\vspace{1mm}
\caption{Segmentation results for the Vaihingen dataset.}
\label{tab:vaihingen}

\begin{tabular}{lcc}
\toprule
Method & Dice & mIoU \\
\midrule
FCN~\cite{badrinarayanan2017segnet} & 28.84 &  28.71 \\
U-Net~\cite{ronneberger2015u} & 79.43 & 65.99 \\
U-Net++~\cite{zhou2018unet++} & 79.49 & 66.04 \\
Res-UNet~\cite{xiao2018weighted} & 79.49 & 66.07 \\
A.A U-Net~\cite{wang2020axial} & 76.83 & 62.49 \\
MedT~\cite{valanarasu2021medical} & 79.55 & 66.17 \\
Ours & {\bf 81.59} & {\bf 69.00} \\
\bottomrule
\end{tabular}
\vspace{1mm}
\caption{Segmentation results for the MoNuSeg dataset.}
\label{tab:Monu}
\vspace{-6mm}
\end{table}

The experiments were carried out with a batch size of eight images, with 12 RRDB blocks. The network depth was seven, and the number of channels in each depth was $[C, C, C, 2C, 2C, 4C, 4C]$, with $C=128$. We used the same augmentation scheme as in~\cite{valanarasu2021medical} with random cropping of $256 \times 256$ to adjust for GPU memory.

It is worth noting that all baseline methods except Segformer and Stdf rely on pre-trained weights obtained on the ImageNet, PASCAL or COCO datasets. Our networks are initialized with random weights.

\noindent{\bf Results\quad}
Following previous work, Cityscapes is evaluated in one of two settings. \textit{Tight}: in this setting, the samples (image and associated segmentation map) are extracted by a tight crop around the object mask. 
\textit{Expansion}: samples are extracted by a crop around the object mask, which is 15\% larger than the tight crop. The inputs of the model are crops 10\% - 20\% larger than the tight one. This setting is slightly more challenging, since there is less information on the location of the target object.

The results for the Cityscapes dataset are reported in Tab.~\ref{tab:cityscapes}. As can be seen, our method outperforms all baseline methods, across all categories and in both settings. 

\begin{figure}[t]
\centering
    \includegraphics[width=1.0\linewidth]{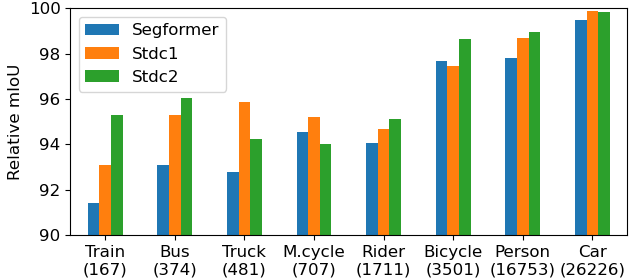}
    \vspace{-.7cm}
    \captionof{figure}{mIoU relative to SegDiff being 100\% for each Cityscape class, sorted by the number of training images per class.}
    \label{fig:models_by_number_of_training_images}
\hfill
\vspace{2.9mm}

 \begin{tabular}{c l c c c c}
 \midrule
 & \multicolumn{1}{l}{Method} & \multicolumn{1}{c}{F1-Score} & \multicolumn{1}{c}{IoU} & 
 \multicolumn{1}{c}{WCov} & \multicolumn{1}{c}{FBound}\\
 \midrule
 \parbox[t]{3mm}{\multirow{7}{*}{\rotatebox[origin=c]{90}{Vaihingen}}} 
& Variant one
& 91.60 & 85.45 & 88.67 & 71.70\\
& Variant two
& 90.92 & 84.00 & 89.67 & 70.00\\
& Variant three
& 93.77 & 88.67 & 91.69 & 80.15 \\
& Variant four
& 94.77 & 90.27 & 93.82 & 82.64\\
& Variant five
& 93.16 & 87.76 & 91.08 & 79.89 \\
& Variant six
& 91.97 & 85.57 & 89.83 & 71.04\\
& Full method
& \textbf{94.95} & \textbf{90.64} & \textbf{94.00} & \textbf{84.37}\\
\midrule
\midrule
 \parbox[t]{3mm}{\multirow{7}{*}{\rotatebox[origin=c]{90}{Cityscapes ``Bus''}}} 
& Variant one
& 90.52 & 84.15 & 90.37 & 62.66\\
& Variant two
& 85.21 & 75.92 & 81.15 & 38.81\\
& Variant three
& 90.35 & 83.76 & 88.56 & 58.80 \\
& Variant four
& \textbf{91.30} & \textbf{85.17} & \textbf{90.34} & 63.85\\
& Variant five
& 89.57 & 82.73 & 88.87 & 58.40 \\
& Variant six
& 82.97 & 72.66 & 80.85 & 34.38\\
& Full method
& 90.72 & 84.35 & 89.96 & \textbf{63.87}\\
\midrule
 \end{tabular}
\captionof{table}{Ablation study for different conditioning methods.}
\label{tab:ablation_conditioning_methods}

\vspace{-6mm}
\end{figure}

The gap is apparent even for the most recent baseline methods and, as can be seen in Fig.~\ref{fig:models_by_number_of_training_images}, the gap in performance is especially sizable for datasets with less training images. 

The results for the Vaihingen dataset are presented in Tab.~\ref{tab:vaihingen}. As can be seen, our method outperforms the results reported in previous work for all four scores.

The results for the MoNuSeg dataset are presented in Tab.~\ref{tab:Monu}. In both segmentation metrics, our method outperforms all previous works, including very recent variants of U-Net and transformers that were developed specifically for this segmentation task.

\begin{figure*}[t]
\centering
\begin{tabular}{@{}c@{~}c@{~}c@{}}
    \includegraphics[width=0.496\linewidth]{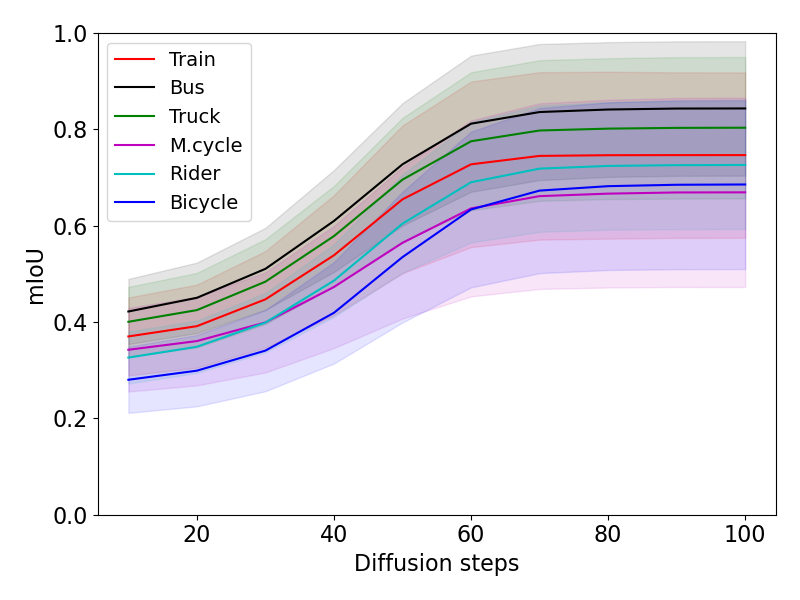} &  
    \includegraphics[width=0.496\linewidth]{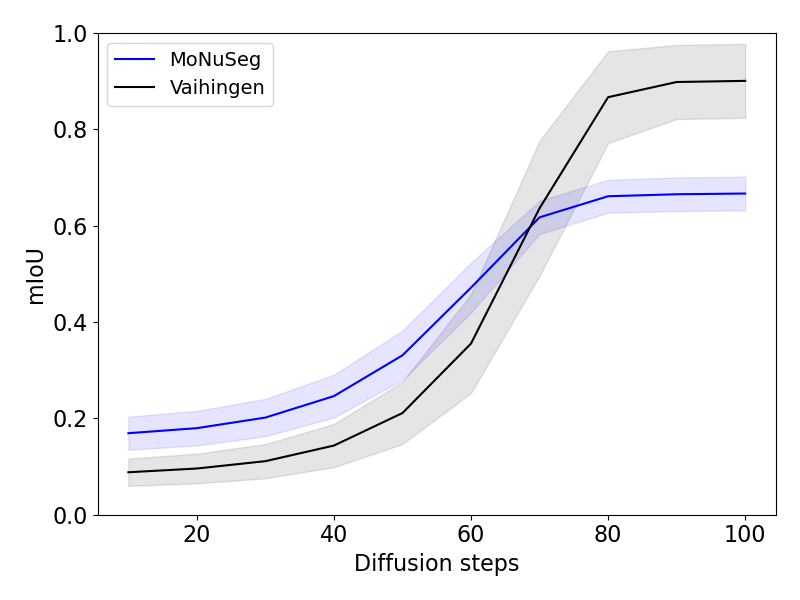} 
    \\
    (a) & (b)\\
\end{tabular}
\caption{mIoU (mean and variance) across the test images as a function of the number of diffusion steps. (a) Results for the Cityscapes classes, with $128\times 128$ image resolution. (b) Results for the Vaihingen and MoNuSeg datasets, with $256\times 256$ image resolution.}
\label{fig:datasets_miou}
\end{figure*}

\begin{figure*}[t]
\centering
\begin{tabular}{@{}c@{~}c@{~}c@{}}
    \includegraphics[width=0.496\linewidth]{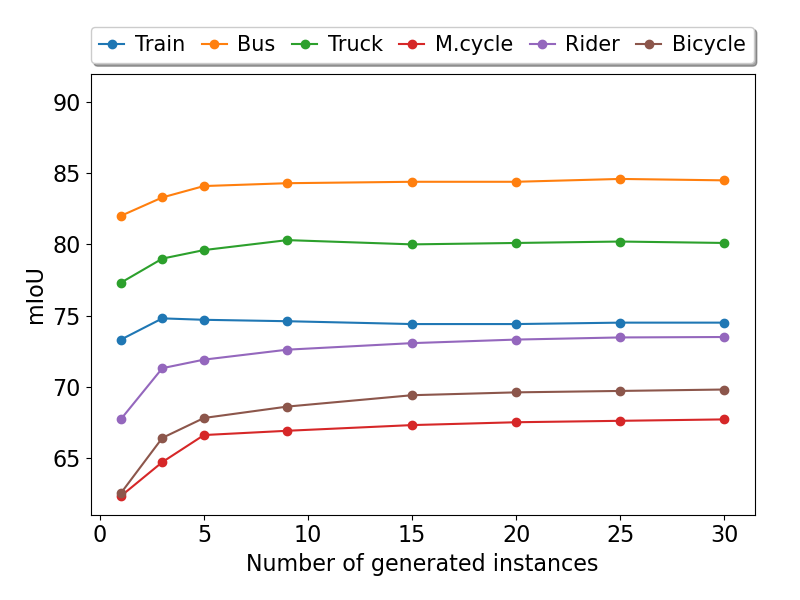} &  
    \includegraphics[width=0.496\linewidth]{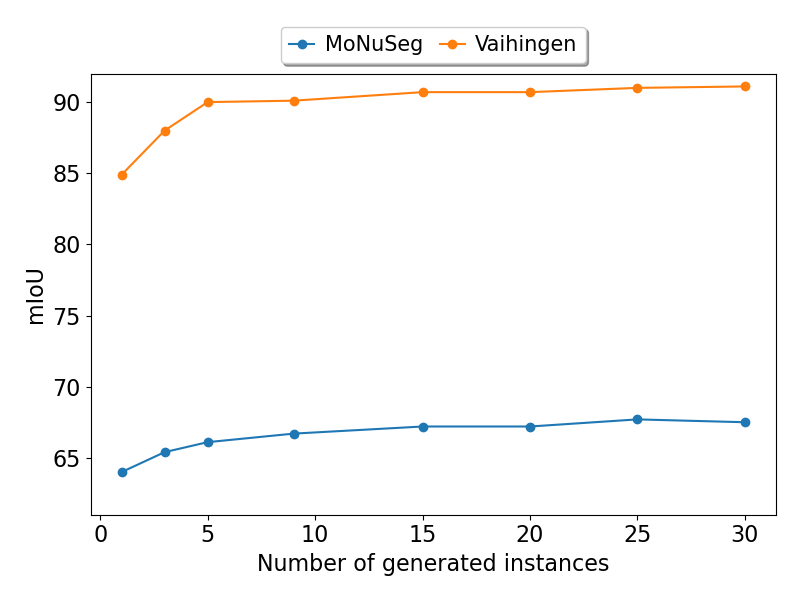} 
    \\
    (a) & (b)\\
\end{tabular}
\caption{mIoU per number of generated inferences. (a) Results for the Cityscapes classes, with $128\times 128$ image resolution. (b) Results for the Vaihingen and MoNuSeg datasets, with $256\times 256$ image resolution.}
\label{fig:datasets_miou_for_number_of_generated_instances}
\end{figure*}

The performance of the mIoU segmentation metric as a function of the number of iterations is presented for the three datasets in Fig.~\ref{fig:datasets_miou}. It is interesting to note that the number of diffusion steps required to achieve the maximal score differs across the datasets. 

All Cityscapes classes present similar behavior (with different levels of performance), saturating around the 60th iteration. The Vaihingen score takes longer to reach its maximal value. While one may attribute this to the larger input image size observed by the network, MoNuSeg, which has the same input image size as Vaihingen, reaches saturation earlier, similarly to the Cityscapes classes. An alternative hypothesis can relate the number of required iterations to the ratio of pixels within the segmentation mask, which is higher for Cityscapes and MoNuSeg than for Vaihingen. This requires further validation. 

\begin{figure*}[t]
\centering
\begin{tabular}{@{}c@{~}c@{~}c@{}}
    \includegraphics[width=0.496\linewidth]{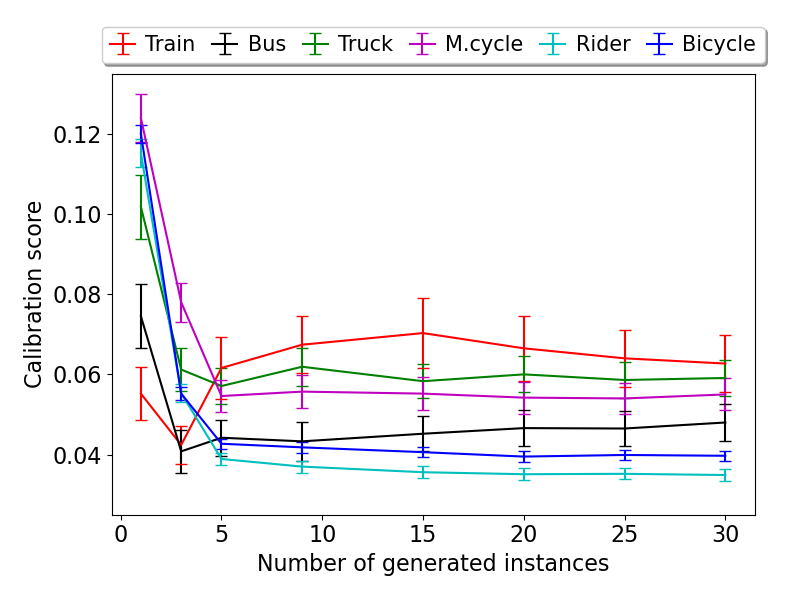} &  
    \includegraphics[width=0.496\linewidth]{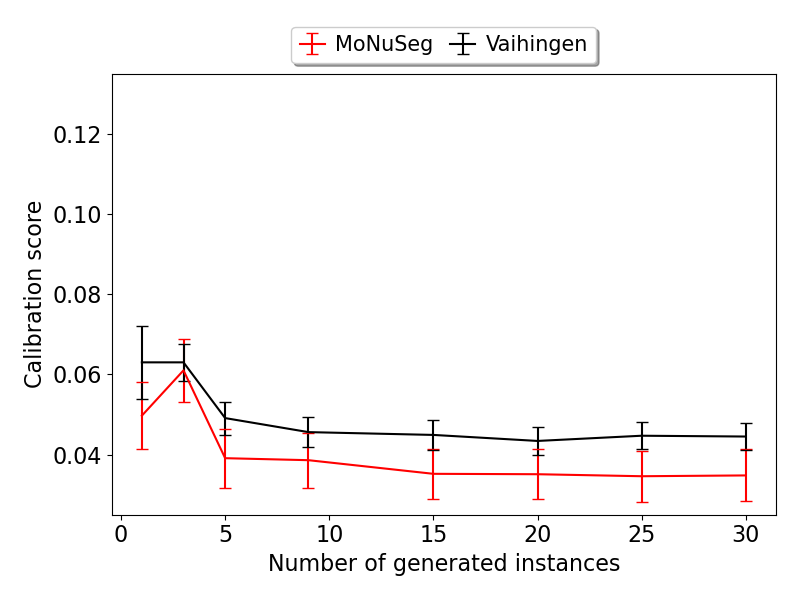} 
    \\
    (a) & (b)\\
\end{tabular}

\caption{Mean calibration score (lower is better) per number of generated inferences. The error bars depict the standard error. (a) Results for the Cityscapes classes, with an image resolution of $128\times 128$. (b) Results for the Vaihingen and MoNuSeg datasets, with the $256\times 256$ image resolution.}
\label{fig:datasets_brier_score_for_number_of_generated_instances}
\end{figure*}

\begin{figure*}[t]
\centering
\begin{tabular}{@{}c@{~}c@{~}c@{~}c@{~}c@{~}c@{~}c@{~}c@{}}
     
    \includegraphics[width=0.13821\textwidth]{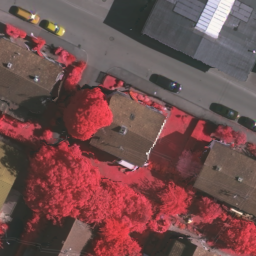} & 
    \includegraphics[width=0.13821\textwidth]{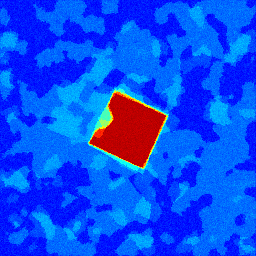} & 
    \includegraphics[width=0.13821\textwidth]{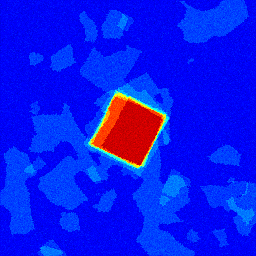} & \includegraphics[width=0.13821\textwidth]{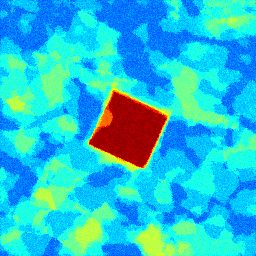} &  
    \includegraphics[width=0.13821\textwidth]{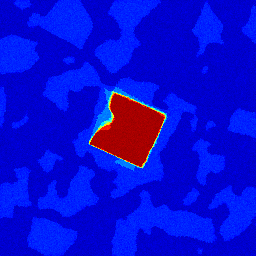} & 
    \includegraphics[width=0.13821\textwidth]{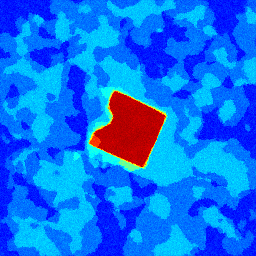}&
    \includegraphics[width=0.13821\textwidth]{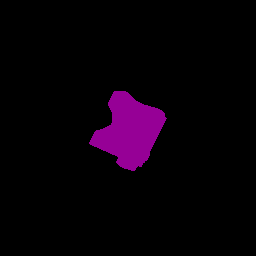}\\
     
    \includegraphics[width=0.13821\textwidth]{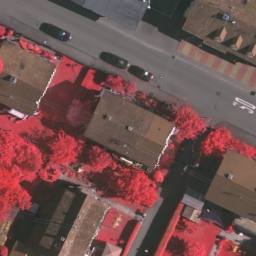} &   \includegraphics[width=0.13821\textwidth]{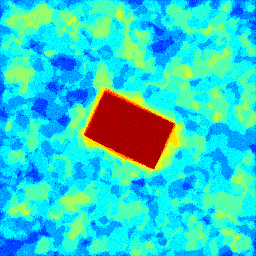} &  
    \includegraphics[width=0.13821\textwidth]{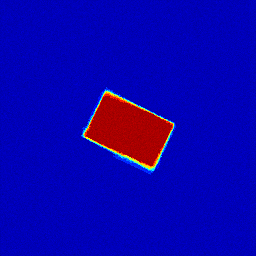} & 
    \includegraphics[width=0.13821\textwidth]{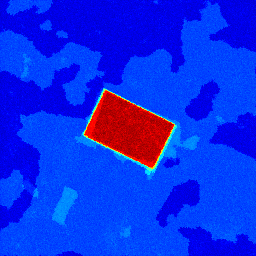} &  
    \includegraphics[width=0.13821\textwidth]{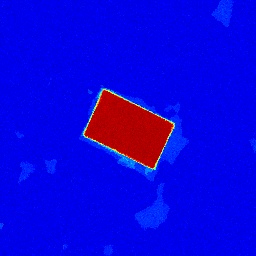} & 
    \includegraphics[width=0.13821\textwidth]{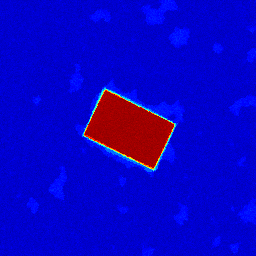} &
    \includegraphics[width=0.13821\textwidth]{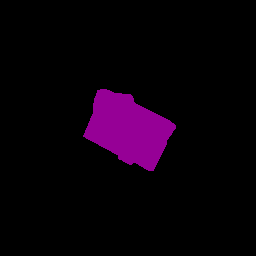}\\

    \includegraphics[width=0.13821\textwidth]{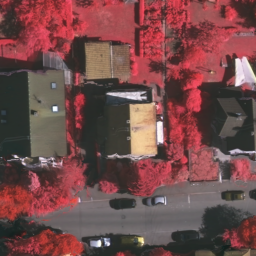} &
    \includegraphics[width=0.13821\textwidth]{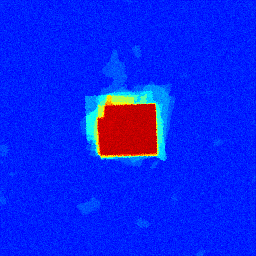} &  
    \includegraphics[width=0.13821\textwidth]{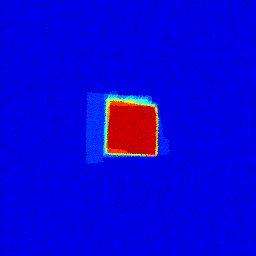} & 
    \includegraphics[width=0.13821\textwidth]{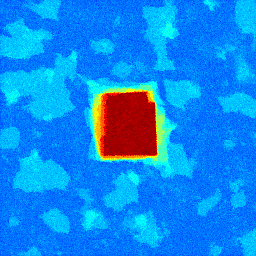} &  
    \includegraphics[width=0.13821\textwidth]{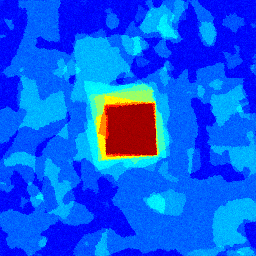} & 
    \includegraphics[width=0.13821\textwidth]{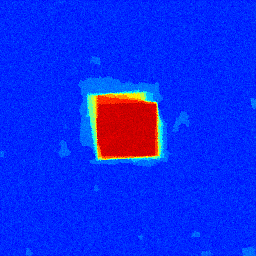}&
     \includegraphics[width=0.13821\textwidth]{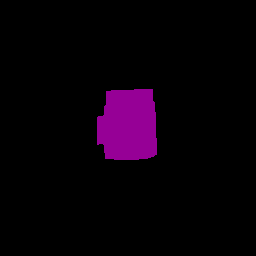}\\

(a) & (b) & (c) & (d) & (e) & (f) & (g)\\
\end{tabular}

\caption{Results of the ablation study. (a) the input image, (b-e) results for variants one--four of our method, respectively, (f) the result of our method, and (g) ground truth.  Panels (b-f) employ the jet color scale between 0 in blue and 1 in red}
\label{fig:ablation_figures}
\end{figure*}

We next study the effect of the number of generated instances on performance. The results can be seen in Fig.~\ref{fig:datasets_miou_for_number_of_generated_instances}. In general, increasing the number of generated instances tends to increase the mIoU score. However, the number of runs required to reach optimal performance varies between classes. For example, for the ``Bus'' and ``Train'' classes of Cityscapes, the best score is achieved when using 10 and 3 generated instances, respectively. MoNuSeg, requires considerably more runs (25) for maximal performance. On the other hand, when the number of generated instances is increased, inference time also increases linearly, resulting in a slower method compared to architectures such as Segformer and Stdc.

Another aspect of achieving improvement by employing multiple generations is calibration. The calibration score is measured as the difference between the prediction probability and the true probability of the event. For example, a perfectly calibrated model is defined by $\mathbb{P}(\hat{Y} = Y | \hat{P} = p) = p$, which means that the prediction probability equals the true probability of the event. We estimate the calibration score by splitting the [0, 1] range into ten uniform bins, then average the squared difference between each bin's mean prediction probability and the percentage of positive samples. 

The results of examining the calibration scores are presented in Fig.~\ref{fig:datasets_brier_score_for_number_of_generated_instances}. For most datasets, increasing the number of generated instances improves the calibration score, especially when the increase is from a single instance. In addition, for the larger classes in Cityscapes - Rider and Bicycle - and for the MoNuSeg and Vaihingen datasets, the improvement continues to increase even more compared to the other datasets. The ``Train'' class in Cityscapes is an exception; here, the single-instance calibration score is better than other experiments with a larger number of generated instances. This phenomenon may be a result of the highly varied size and the small number of test images. 

\begin{figure*}[t]
\centering
\begin{tabular}{@{}c@{~}c@{~}c@{}}
    \includegraphics[width=0.49321025\linewidth]{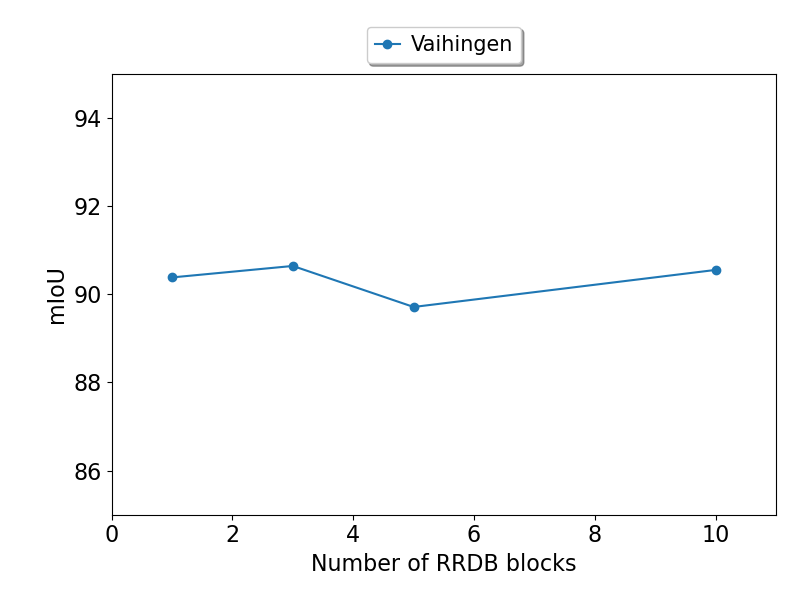} & 
    \includegraphics[width=0.49321025\linewidth]{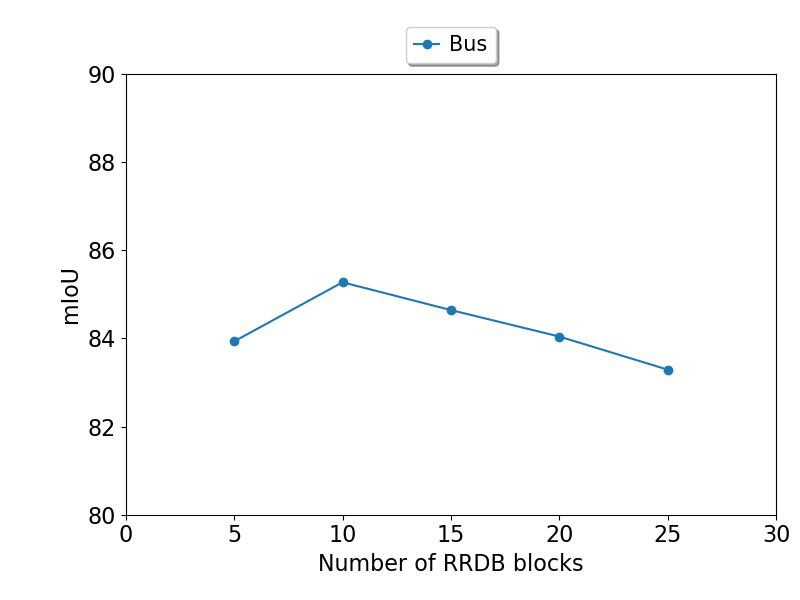} \\
    (a) & (b)
\end{tabular}

\caption{mIoU per number of RRDB blocks. (a) Results on Vaihingen, (b) Results on Cityscapes ``Bus''.}

\label{fig:iou_per_rrdb_blocks}
\end{figure*}

\begin{figure*}[t]
\centering
\begin{tabular}{@{}c@{~}c@{~}c@{}}
    \includegraphics[width=0.4921025\linewidth]{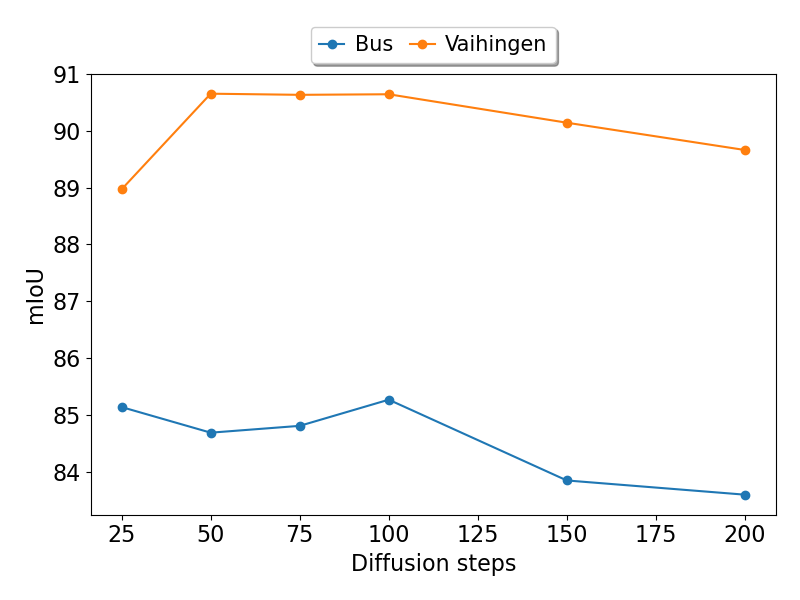} & 
    \includegraphics[width=0.4921025\linewidth]{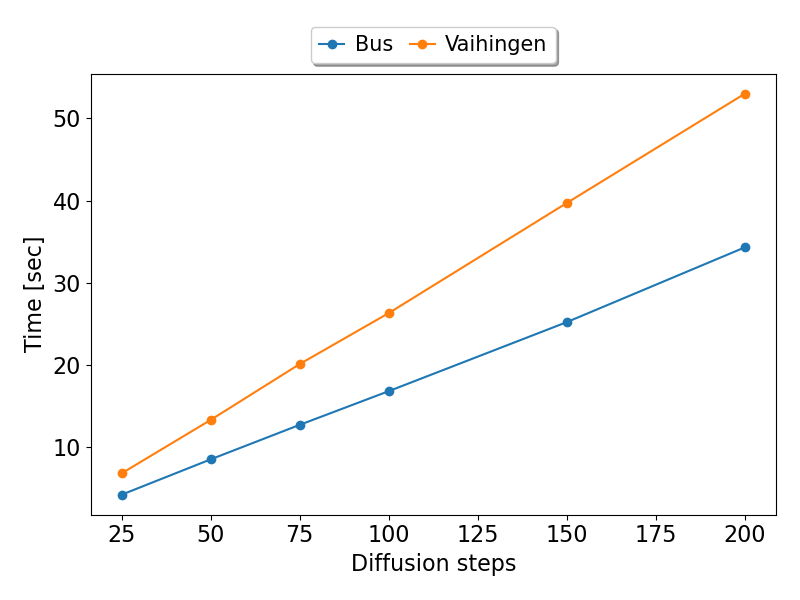} 
    \\
    (a) & (b)
\end{tabular}
\caption{Generation time in seconds and mIou per number of diffusion steps for Vaihingen and Cityscapes ``Bus''. (a) mIoU per diffusion step, (b) Time per diffusion step.}
\label{fig:time_and_iou_per_steps}
\end{figure*}

\noindent{\bf Ablation Study\quad}
We evaluate various alternatives to our method. The first variant concatenates $[F(x_t),G(I)]$ at the channel dimension. The second variant employs FC-HarDNet-70 V2~\cite{chao2019hardnet} instead of RRDBs.
The third variant, following~\cite{saharia2021image,ho2022cascaded}, concatenates $I$ channelwise to $x_t$, without using an encoder. The last alternative method is to propagate $F(x_t)$ through the U-Net module and add it to $G(I)$ after the first, third, and fifth downsample blocks (variants four--six), instead of performing $F(x_t) + G(I)$. In this variant, $G(I)$ is downsampled to match the required number of channels by propagating it through a 2D-convolutional layer with a stride of two.

These variant experiments were tested by averaging nine generated instances on the Vaihingen dataset and on Cityscapes ``Bus'' (the performance reported for our method is therefore slightly different from those reported in Tab.~\ref{tab:vaihingen}).

The summation we introduce as a conditioning approach outperforms concatenation (variant one) on Vaihingen by a large margin, while on Cityscapes ``Bus'', the difference is small. The RRDB blocks are preferable to the FC-HarDNet architecture in both datasets (variant two). Removing the encoder affects the metrics significantly (variant three), slightly more so on Vaihingen. The change in the signal's integration position of variant four leads to a negligible difference on Vaihingen and even outperforms our full method on Cityscapes ``Bus''. Variants five and six lead to a decrease in performance as the distance from the first layer increases. Fig.~\ref{fig:ablation_figures} depicts sample results for the various variants (one--four) of the Vaihingen dataset. 

\noindent{\bf Parameter sensitivity\quad} For testing the stability of our proposed method, we experimented with the two hyperparameters that can affect performance the most: the number of diffusion steps, and the number of RRDB blocks. To study the effect of these parameters, we varied the number of diffusion steps in the range of [25, 50, 75, 100, 150, 200], and the number of RRDB blocks in the range of [1, 3, 5, 10] for Vaihingen and [5, 10, 15, 20, 25] for Cityscapes ``Bus''. We started from a baseline configuration (which was 100 diffusion steps, 3 RRDB blocks for Vaihingen, and 10 RRDB blocks for Cityscapes ``Bus'') and experimented with different values around these. 

\noindent{\bf The effect of the number of RRDB blocks\quad}
In this part we set the number of diffusion steps to 100. As can be seen in Fig.~\ref{fig:iou_per_rrdb_blocks}, with our configuration, the optimal number of RRDB blocks is 3 for Vaihingen, and 10 for Cityscapes ``Bus''. However, evidently, the number of blocks has a limited impact in the case of both Cityscapes and Vaihingen. The gap between the best and worst performance points is less than 1 mIoU for Vaihingen and less than 2 mIoU for Cityscapes ``Bus''. Therefore, we conclude that this hyperparameter has a small effect on performance.

\noindent{\bf Varying the number of diffusion steps $T$\quad}\label{subsec:varying_diffusion_steps}
In this part, we set the number of RRDB blocks of Vaihingen to 3 and Cityscapes ``Bus'' to 10.
We explore the possible accuracy/runtime tradeoff with regards to the number $T$ of diffusion steps. Results are shown in Fig.~\ref{fig:time_and_iou_per_steps}.

When the number of diffusion steps is increased - as we can see in Fig.~\ref{fig:time_and_iou_per_steps}(a) - the graph fluctuation for Vaihingen is less than 1 mIoU, and for Cityscapes ``Bus'' it is less than 2 mIoU.

Surprisingly, when the number of diffusion steps is reduced, even to just 25, which is a very low number compared to the literature~\cite{ho2020denoising,nichol2021improved}, the segmentation results remain stable in both datasets, with a degradation of only up to 2 mIou for Vaihingen, and 1 mIou for Cityscapes ``Bus''. This reduction can speed up performance by a factor of four and provide a reasonable accuracy to runtime tradeoff.

The results for the generation time of one sample in seconds are presented in Fig.~\ref{fig:time_and_iou_per_steps}(b). As can be observed, both graphs are linear, with a different slope. The main reason for this is the difference in image size (which is $256 \times 256$ for Vaihingen and $128 \times 128$ for Cityscapes ``Bus''). Another, minor reason is the difference in the number of RRDB blocks in this experiment.

\section{Conclusions}

A wealth of methods have been applied to image segmentation, including active contour and their deep variants, encoder-decoder architectures, and U-Nets, which - together with more recent, transformer-based methods - represent a leading approach. In this work, we propose utilizing the state-of-the-art image generation technique of diffusion models. Our diffusion model employs a U-Net architecture, which is used to incrementally improve the obtained generation, similarly to other recent diffusion models. 

In order to condition the input image, we generate another encoding path, which is similar to U-Net's encoder-decoder use in conventional image segmentation methods. The two encoder pathways are merged by summing the activations early in the U-Net's encoder. 

Using our approach, we obtain state-of-the-art segmentation results on a diverse set of benchmarks, including street view images, aerial images, and microscopy.

\section{Acknowledgments}
This project has received funding from the European Research Council (ERC) under the European Unions Horizon 2020 research and innovation programme (grant ERC CoG 725974). 

{\small
\bibliographystyle{ieee_fullname}
\bibliography{seg}

\begin{thebibliography}{10}\itemsep=-1pt

\bibitem{acuna2018efficient}
David Acuna, Huan Ling, Amlan Kar, and Sanja Fidler.
\newblock Efficient interactive annotation of segmentation datasets with
  polygon-rnn++.
\newblock In {\em Proceedings of the IEEE conference on Computer Vision and
  Pattern Recognition}, pages 859--868, 2018.

\bibitem{austin2021structured}
Jacob Austin, Daniel~D Johnson, Jonathan Ho, Daniel Tarlow, and Rianne van~den
  Berg.
\newblock Structured denoising diffusion models in discrete state-spaces.
\newblock {\em Advances in Neural Information Processing Systems},
  34:17981--17993, 2021.

\bibitem{badrinarayanan2017segnet}
Vijay Badrinarayanan, Alex Kendall, and Roberto Cipolla.
\newblock Segnet: A deep convolutional encoder-decoder architecture for image
  segmentation.
\newblock {\em IEEE transactions on pattern analysis and machine intelligence},
  39(12):2481--2495, 2017.

\bibitem{chao2019hardnet}
Ping Chao, Chao-Yang Kao, Yu-Shan Ruan, Chien-Hsiang Huang, and Youn-Long Lin.
\newblock Hardnet: A low memory traffic network.
\newblock In {\em Proceedings of the IEEE/CVF international conference on
  computer vision}, pages 3552--3561, 2019.

\bibitem{chen2017deeplab}
Liang-Chieh Chen, George Papandreou, Iasonas Kokkinos, Kevin Murphy, and Alan~L
  Yuille.
\newblock Deeplab: Semantic image segmentation with deep convolutional nets,
  atrous convolution, and fully connected crfs.
\newblock {\em IEEE transactions on pattern analysis and machine intelligence},
  40(4):834--848, 2017.

\bibitem{chen2020wavegrad}
Nanxin Chen, Yu Zhang, Heiga Zen, Ron~J Weiss, Mohammad Norouzi, and William
  Chan.
\newblock Wavegrad: Estimating gradients for waveform generation.
\newblock {\em arXiv preprint arXiv:2009.00713}, 2020.

\bibitem{cheng2019darnet}
Dominic Cheng, Renjie Liao, Sanja Fidler, and Raquel Urtasun.
\newblock Darnet: Deep active ray network for building segmentation.
\newblock In {\em Proceedings of the IEEE/CVF Conference on Computer Vision and
  Pattern Recognition}, pages 7431--7439, 2019.

\bibitem{choi2021ilvr}
Jooyoung Choi, Sungwon Kim, Yonghyun Jeong, Youngjune Gwon, and Sungroh Yoon.
\newblock Ilvr: Conditioning method for denoising diffusion probabilistic
  models.
\newblock {\em arXiv preprint arXiv:2108.02938}, 2021.

\bibitem{Cordts2016Cityscapes}
Marius Cordts, Mohamed Omran, Sebastian Ramos, Timo Rehfeld, Markus Enzweiler,
  Rodrigo Benenson, Uwe Franke, Stefan Roth, and Bernt Schiele.
\newblock The cityscapes dataset for semantic urban scene understanding.
\newblock In {\em Proceedings of the IEEE conference on computer vision and
  pattern recognition}, pages 3213--3223, 2016.

\bibitem{dhariwal2021diffusion}
Prafulla Dhariwal and Alexander Nichol.
\newblock Diffusion models beat gans on image synthesis.
\newblock {\em Advances in Neural Information Processing Systems}, 34, 2021.

\bibitem{fan2021rethinking}
Mingyuan Fan, Shenqi Lai, Junshi Huang, Xiaoming Wei, Zhenhua Chai, Junfeng
  Luo, and Xiaolin Wei.
\newblock Rethinking bisenet for real-time semantic segmentation.
\newblock In {\em Proceedings of the IEEE/CVF conference on computer vision and
  pattern recognition}, pages 9716--9725, 2021.

\bibitem{fischer2017adversarial}
Volker Fischer, Mummadi~Chaithanya Kumar, Jan~Hendrik Metzen, and Thomas Brox.
\newblock Adversarial examples for semantic image segmentation.
\newblock {\em arXiv preprint arXiv:1703.01101}, 2017.

\bibitem{fu2020scene}
Jun Fu, Jing Liu, Jie Jiang, Yong Li, Yongjun Bao, and Hanqing Lu.
\newblock Scene segmentation with dual relation-aware attention network.
\newblock {\em IEEE Transactions on Neural Networks and Learning Systems},
  32(6):2547--2560, 2020.

\bibitem{gur2019end}
Shir Gur, Tal Shaharabany, and Lior Wolf.
\newblock End to end trainable active contours via differentiable rendering.
\newblock {\em arXiv preprint arXiv:1912.00367}, 2019.

\bibitem{hatamizadeh2020end}
Ali Hatamizadeh, Debleena Sengupta, and Demetri Terzopoulos.
\newblock End-to-end trainable deep active contour models for automated image
  segmentation: Delineating buildings in aerial imagery.
\newblock In {\em European Conference on Computer Vision}, pages 730--746.
  Springer, 2020.

\bibitem{he2017mask}
Kaiming He, Georgia Gkioxari, Piotr Doll{\'a}r, and Ross Girshick.
\newblock Mask r-cnn.
\newblock In {\em Proceedings of the IEEE international conference on computer
  vision}, pages 2961--2969, 2017.

\bibitem{he2016deep}
Kaiming He, Xiangyu Zhang, Shaoqing Ren, and Jian Sun.
\newblock Deep residual learning for image recognition.
\newblock In {\em Proceedings of the IEEE conference on computer vision and
  pattern recognition}, pages 770--778, 2016.

\bibitem{ho2020denoising}
Jonathan Ho, Ajay Jain, and Pieter Abbeel.
\newblock Denoising diffusion probabilistic models.
\newblock {\em Advances in Neural Information Processing Systems},
  33:6840--6851, 2020.

\bibitem{ho2022cascaded}
Jonathan Ho, Chitwan Saharia, William Chan, David~J Fleet, Mohammad Norouzi,
  and Tim Salimans.
\newblock Cascaded diffusion models for high fidelity image generation.
\newblock {\em Journal of Machine Learning Research}, 23(47):1--33, 2022.

\bibitem{hoogeboom2021argmax}
Emiel Hoogeboom, Didrik Nielsen, Priyank Jaini, Patrick Forr{\'e}, and Max
  Welling.
\newblock Argmax flows and multinomial diffusion: Towards non-autoregressive
  language models.
\newblock {\em arXiv e-prints}, pages arXiv--2102, 2021.

\bibitem{huang2021variational}
Chin-Wei Huang, Jae~Hyun Lim, and Aaron~C Courville.
\newblock A variational perspective on diffusion-based generative models and
  score matching.
\newblock {\em Advances in Neural Information Processing Systems}, 34, 2021.

\bibitem{huang2019ccnet}
Zilong Huang, Xinggang Wang, Lichao Huang, Chang Huang, Yunchao Wei, and Wenyu
  Liu.
\newblock Ccnet: Criss-cross attention for semantic segmentation.
\newblock In {\em Proceedings of the IEEE/CVF International Conference on
  Computer Vision}, pages 603--612, 2019.

\bibitem{kingma2021variational}
Diederik~P Kingma, Tim Salimans, Ben Poole, and Jonathan Ho.
\newblock Variational diffusion models.
\newblock {\em arXiv preprint arXiv:2107.00630}, 2021.

\bibitem{kong2020diffwave}
Zhifeng Kong, Wei Ping, Jiaji Huang, Kexin Zhao, and Bryan Catanzaro.
\newblock Diffwave: A versatile diffusion model for audio synthesis.
\newblock {\em arXiv preprint arXiv:2009.09761}, 2020.

\bibitem{kumar2019multi}
Neeraj Kumar, Ruchika Verma, Deepak Anand, Yanning Zhou, Omer~Fahri Onder,
  Efstratios Tsougenis, Hao Chen, Pheng-Ann Heng, Jiahui Li, Zhiqiang Hu,
  et~al.
\newblock A multi-organ nucleus segmentation challenge.
\newblock {\em IEEE transactions on medical imaging}, 39(5):1380--1391, 2019.

\bibitem{kumar2017dataset}
Neeraj Kumar, Ruchika Verma, Sanuj Sharma, Surabhi Bhargava, Abhishek Vahadane,
  and Amit Sethi.
\newblock A dataset and a technique for generalized nuclear segmentation for
  computational pathology.
\newblock {\em IEEE transactions on medical imaging}, 36(7):1550--1560, 2017.

\bibitem{li2022srdiff}
Haoying Li, Yifan Yang, Meng Chang, Shiqi Chen, Huajun Feng, Zhihai Xu, Qi Li,
  and Yueting Chen.
\newblock Srdiff: Single image super-resolution with diffusion probabilistic
  models.
\newblock {\em Neurocomputing}, 2022.

\bibitem{liang2020polytransform}
Justin Liang, Namdar Homayounfar, Wei-Chiu Ma, Yuwen Xiong, Rui Hu, and Raquel
  Urtasun.
\newblock Polytransform: Deep polygon transformer for instance segmentation.
\newblock In {\em Proceedings of the IEEE/CVF Conference on Computer Vision and
  Pattern Recognition}, pages 9131--9140, 2020.

\bibitem{ling2019fast}
Huan Ling, Jun Gao, Amlan Kar, Wenzheng Chen, and Sanja Fidler.
\newblock Fast interactive object annotation with curve-gcn.
\newblock In {\em Proceedings of the IEEE/CVF Conference on Computer Vision and
  Pattern Recognition}, pages 5257--5266, 2019.

\bibitem{liu2021diffsinger}
Jinglin Liu, Chengxi Li, Yi Ren, Feiyang Chen, Peng Liu, and Zhou Zhao.
\newblock Diffsinger: Singing voice synthesis via shallow diffusion mechanism.
\newblock {\em arXiv preprint arXiv:2105.02446}, 2021.

\bibitem{long2015fully}
Jonathan Long, Evan Shelhamer, and Trevor Darrell.
\newblock Fully convolutional networks for semantic segmentation.
\newblock In {\em Proceedings of the IEEE conference on computer vision and
  pattern recognition}, pages 3431--3440, 2015.

\bibitem{loshchilov2017decoupled}
Ilya Loshchilov and Frank Hutter.
\newblock Decoupled weight decay regularization.
\newblock {\em arXiv preprint arXiv:1711.05101}, 2017.

\bibitem{luc2016semantic}
Pauline Luc, Camille Couprie, Soumith Chintala, and Jakob Verbeek.
\newblock Semantic segmentation using adversarial networks.
\newblock {\em arXiv preprint arXiv:1611.08408}, 2016.

\bibitem{marcos2018learning}
Diego Marcos, Devis Tuia, Benjamin Kellenberger, Lisa Zhang, Min Bai, Renjie
  Liao, and Raquel Urtasun.
\newblock Learning deep structured active contours end-to-end.
\newblock In {\em Proceedings of the IEEE Conference on Computer Vision and
  Pattern Recognition}, pages 8877--8885, 2018.

\bibitem{nichol2021improved}
Alexander~Quinn Nichol and Prafulla Dhariwal.
\newblock Improved denoising diffusion probabilistic models.
\newblock In {\em International Conference on Machine Learning}, pages
  8162--8171. PMLR, 2021.

\bibitem{nirkin2021hyperseg}
Yuval Nirkin, Lior Wolf, and Tal Hassner.
\newblock Hyperseg: Patch-wise hypernetwork for real-time semantic
  segmentation.
\newblock In {\em Proceedings of the IEEE/CVF Conference on Computer Vision and
  Pattern Recognition}, pages 4061--4070, 2021.

\bibitem{popov2021grad}
Vadim Popov, Ivan Vovk, Vladimir Gogoryan, Tasnima Sadekova, and Mikhail
  Kudinov.
\newblock Grad-tts: A diffusion probabilistic model for text-to-speech.
\newblock In {\em International Conference on Machine Learning}, pages
  8599--8608. PMLR, 2021.

\bibitem{ronneberger2015u}
Olaf Ronneberger, Philipp Fischer, and Thomas Brox.
\newblock U-net: Convolutional networks for biomedical image segmentation.
\newblock In {\em International Conference on Medical image computing and
  computer-assisted intervention}, pages 234--241. Springer, 2015.

\bibitem{rottensteiner2014isprs}
Franz Rottensteiner, Gunho Sohn, Markus Gerke, and Jan~D Wegner.
\newblock Isprs semantic labeling contest.
\newblock {\em ISPRS: Leopoldsh{\"o}he, Germany}, 2014.

\bibitem{saharia2021palette}
Chitwan Saharia, William Chan, Huiwen Chang, Chris~A Lee, Jonathan Ho, Tim
  Salimans, David~J Fleet, and Mohammad Norouzi.
\newblock Palette: Image-to-image diffusion models.
\newblock {\em arXiv preprint arXiv:2111.05826}, 2021.

\bibitem{saharia2021image}
Chitwan Saharia, Jonathan Ho, William Chan, Tim Salimans, David~J Fleet, and
  Mohammad Norouzi.
\newblock Image super-resolution via iterative refinement.
\newblock {\em arXiv preprint arXiv:2104.07636}, 2021.

\bibitem{san2021noise}
Robin San-Roman, Eliya Nachmani, and Lior Wolf.
\newblock Noise estimation for generative diffusion models.
\newblock {\em arXiv preprint arXiv:2104.02600}, 2021.

\bibitem{sohl2015deep}
Jascha Sohl-Dickstein, Eric Weiss, Niru Maheswaranathan, and Surya Ganguli.
\newblock Deep unsupervised learning using nonequilibrium thermodynamics.
\newblock In {\em International Conference on Machine Learning}, pages
  2256--2265. PMLR, 2015.

\bibitem{strudel2021segmenter}
Robin Strudel, Ricardo Garcia, Ivan Laptev, and Cordelia Schmid.
\newblock Segmenter: Transformer for semantic segmentation.
\newblock In {\em Proceedings of the IEEE/CVF International Conference on
  Computer Vision}, pages 7262--7272, 2021.

\bibitem{valanarasu2021medical}
Jeya Maria~Jose Valanarasu, Poojan Oza, Ilker Hacihaliloglu, and Vishal~M
  Patel.
\newblock Medical transformer: Gated axial-attention for medical image
  segmentation.
\newblock In {\em International Conference on Medical Image Computing and
  Computer-Assisted Intervention}, pages 36--46. Springer, 2021.

\bibitem{wang2020axial}
Huiyu Wang, Yukun Zhu, Bradley Green, Hartwig Adam, Alan Yuille, and
  Liang-Chieh Chen.
\newblock Axial-deeplab: Stand-alone axial-attention for panoptic segmentation.
\newblock In {\em European Conference on Computer Vision}, pages 108--126.
  Springer, 2020.

\bibitem{wang2018esrgan}
Xintao Wang, Ke Yu, Shixiang Wu, Jinjin Gu, Yihao Liu, Chao Dong, Yu Qiao, and
  Chen Change~Loy.
\newblock Esrgan: Enhanced super-resolution generative adversarial networks.
\newblock In {\em Proceedings of the European conference on computer vision
  (ECCV) workshops}, pages 0--0, 2018.

\bibitem{xiao2018weighted}
Xiao Xiao, Shen Lian, Zhiming Luo, and Shaozi Li.
\newblock Weighted res-unet for high-quality retina vessel segmentation.
\newblock In {\em 2018 9th international conference on information technology
  in medicine and education (ITME)}, pages 327--331. IEEE, 2018.

\bibitem{xie2017adversarial}
Cihang Xie, Jianyu Wang, Zhishuai Zhang, Yuyin Zhou, Lingxi Xie, and Alan
  Yuille.
\newblock Adversarial examples for semantic segmentation and object detection.
\newblock In {\em Proceedings of the IEEE international conference on computer
  vision}, pages 1369--1378, 2017.

\bibitem{xie2021segformer}
Enze Xie, Wenhai Wang, Zhiding Yu, Anima Anandkumar, Jose~M Alvarez, and Ping
  Luo.
\newblock Segformer: Simple and efficient design for semantic segmentation with
  transformers.
\newblock {\em Advances in Neural Information Processing Systems}, 34, 2021.

\bibitem{xue2018segan}
Yuan Xue, Tao Xu, Han Zhang, L~Rodney Long, and Xiaolei Huang.
\newblock Segan: adversarial network with multi-scale l1 loss for medical image
  segmentation.
\newblock {\em Neuroinformatics}, 16(3):383--392, 2018.

\bibitem{yu2015multi}
Fisher Yu and Vladlen Koltun.
\newblock Multi-scale context aggregation by dilated convolutions.
\newblock {\em arXiv preprint arXiv:1511.07122}, 2015.

\bibitem{zhou2018unet++}
Zongwei Zhou, Md~Mahfuzur Rahman~Siddiquee, Nima Tajbakhsh, and Jianming Liang.
\newblock Unet++: A nested u-net architecture for medical image segmentation.
\newblock In {\em Deep learning in medical image analysis and multimodal
  learning for clinical decision support}, pages 3--11. Springer, 2018.

\end{thebibliography}
}

\end{document}